\documentclass[letterpaper, 10 pt, conference]{ieeeconf}  % Comment this line out if you need a4paper

\IEEEoverridecommandlockouts                              % This command is only needed if 
\usepackage{graphics} % for pdf, bitmapped graphics files
\usepackage{epsfig} % for postscript graphics files
\usepackage{mathptmx} % assumes new font selection scheme installed
\usepackage{times} % assumes new font selection scheme installed
\usepackage{amsmath} % assumes amsmath package installed
\usepackage{amssymb}  % assumes amsmath package installed
\usepackage{amsfonts}
\usepackage{graphicx}
\usepackage{multirow}
\usepackage{tikz}
\usepackage{cite}
\usepackage{romannum}
\usepackage{empheq}
\usepackage{bm}
\DeclareMathAlphabet{\pazocal}{OMS}{zplm}{m}{n}
\newcommand{\norm}[1]{\left\lVert#1\right\rVert}

\title{\LARGE \bf
A Hybrid Position/Force Controller for Joint Robots
}

\author{Shengwen Xie$^{1}$ and Juan Ren$^{1}$ % <-this % stops a space
\thanks{*This work has been published in the proceedings of ICRA 2021 and we have added some extra explanations in this version.}% <-this % stops a space
\thanks{$^{1}$ Department of Mechanical Engineering,
        Iowa State University, Ames, IA 50011, USA  {\tt\small swxie@outlook.com,}  {\tt\small juanren@iastate.edu}  }
}

\begin{document}

\maketitle
\thispagestyle{empty}
\pagestyle{empty}
\setlength{\textfloatsep}{-0.1cm}
\setlength{\floatsep}{-0.1cm}

%%%%%%%%%%%%%%%%%%%%%%%%%%%%%%%%%%%%%%%%%%%%%%%%%%%%%%%%%%%%%%%%%%%%%%%%%%%%%%%%
\begin{abstract}
In this paper, we present a hybrid position/force controller for operating joint robots. The hybrid controller has two goals---motion tracking and force regulating. As long as these two goals are not mutually exclusive, they can be decoupled in some way. In this work, we make use of the smooth and invertible mapping from the joint space to the task space to decouple the two control goals and design controllers separately. The traditional motion controller in the task space is used for motion control, while the force controller is designed through manipulating the desired trajectory to regulate the force indirectly. Two case studies---contour tracking/polishing surfaces and grabbing boxes with two robotic arms---are presented to show the efficacy of the hybrid controller, and simulations with physics engines are carried out to validate the efficacy of the proposed method.
\end{abstract}

%%%%%%%%%%%%%%%%%%%%%%%%%%%%%%%%%%%%%%%%%%%%%%%%%%%%%%%%%%%%%%%%%%%%%%%%%%%%%%%%
\section{Introduction}
%importance of hybrid controller
In many automation tasks in which the robot interacts with the environment, both position and force of the end-effector need to be controlled to complete the task, i.e., a hybrid motion/force controller is needed. For example, when using a robotic arm to polish the car apron \cite{schindlbeck2015unified}, both the position and contact force should be regulated. 

Controllers for joint robots with one control goal, e.g., force regulation or position control, have been studied extensively. Position controller can be designed in the joint space or task space \cite{lynch2017modern,murray1994mathematical}. Designing force controller is more challenging due to the interaction with the environment and the interaction model is hard to determine. Impedance controllers and admittance controllers can establish a mass-spring-damper model between the end-effector and the environment, but force cannot be accurately controlled\cite{hogan1985impedance,ott2010unified}. By incorporating force feedback in the position controller, accurate force tracking can be achieved \cite{lange2013force,kroger2004adaptive}. When force sensor is not available, sensorless approaches were developed \cite{dehghan2015adaptive}. Although various position or force controllers are available, the choices for hybrid position/force controller are still limited, and designing robust hybrid controller remains an active research area.

One popular position/force controller is proposed in \cite{raibert1981hybrid}, in which the motion and force controllers are applied in two complementary subspaces thus the stability analysis of each controller can be inherited. However, the controller is not robust since the constraint is hard to guarantee in the entire process (e.g., contact loss can happen), and the orthogonality relies on the choice of coordinate and metric \cite{duffy1990fallacy,lynch2017modern}. Marin et al. proposed to use kinestatic filtering to overcome the limitation, but it still relies on the constraints to hold, which makes the controller less robust for applications without natural constraints\cite{marin2016unified}. Recently, the concept of energy tank based on passivity has been used in robot controller design \cite{duindam2004port,schindlbeck2015unified,landi2018passivity}. For example, a Cartesian force/impedance control is developed in \cite{schindlbeck2015unified}. However, the controller is not a position/force controller in the sense that the path to the setpoint cannot be regulated. Iterative learning control (ILC) has been proposed to increase the controller robustness when the nonlinear terms cannot be accurately compensated, but ILC achieves the control  results through repetitions\cite{chen2020robotic}. To deal with the lag of robot position control, Liu et al. proposed to control the position and force with different control frequencies \cite{liu2020frequency}. Although the robustness has been improved, the bandwidth of the position controller is limited. Therefore, robust hybrid position/force controllers which can be used without natural constraints are still not available. 

In this work, we present a hybrid position/force controller which does not rely on the constraints and can regulate both the position and force applied by the end-effector. It is assumed that the two control objectives do not conflict, that is, achieving one objective does not prohibit achieving another. Generally, there are 6 degree of freedoms (DOFs) in the task space. If a smooth and invertible mapping  exists such that the interaction force depends on $n$ DOFs of the 6 DOFs, one can design force controller for the $n$ DOFs and motion controller for the rest $6-n$ DOFs. The motion controller which utilizes a smooth and invertible mapping to transform the configuration from the joint space to the task space is used \cite{murray1994mathematical}. Then, a force controller is incorporated through manipulating the desired position corresponding to the $n$ DOFs. The closed-loop stability of the hybrid controller is analyzed with the linear interaction model between the end-effector and the environment.

To verify the effectiveness of the hybrid controller, two cases studies are presented. The first case is inspired by the task in \cite{schindlbeck2015unified} which can be generalized to a contour tracking problem with a desired contact force. Compared to the approach in \cite{schindlbeck2015unified}, the proposed controller is advantageous in following the pre-designed trajectory while regulating the force in the process. In the second case study, the direction of the force to be controlled can be varying, but it always corresponds to 1 DOF among the 6 DOFs of the end-effector in the task space. The key is to modify the smooth and invertible mapping from the joint space to the task space to ensure that the force controller is related with 1 DOF and thus the proposed hybrid controller framework can be adopted. Although in all the case studies, the force controller always corresponds to 1 DOF, it can be easily extended to more DOFs. Moreover, the proposed controller can be used for mounting a screwdriver onto a screw as reported in \cite{mounting2020}.

\section{Hybrid Controller Design}
Let $R\in SO(3)$ denote the orientation of the end-effector and we can map $R$ to $\Theta=[\Theta_x, \Theta_y, \Theta_z] \in so(3)$\cite{lynch2017modern} (p. 83). Thus, the end-effector pose can be represented with $[\Theta_x, \Theta_y, \Theta_z,x, y, z]$ where $[x, y, z]$ is the position. The hybrid controller design in this work is based on the assumption that the interaction force in some direction can only be affected by the position in that direction. For example, the torque around $x$ axis can only be changed through varying $\Theta_x$ and the interaction force along $x$ axis can be changed by manipulating position in $x$ direction. Note that the interaction between the end-effector and the environment is very complex, the assumption here is to simplify the theoretical analysis and the hybrid controller may still work even when they are violated due to its robustness.

\subsection{Motion Control in the Task Space}
We consider a robotic arm with 6 DOFs. The equation of motion is
\begin{equation}
M(\bm \theta)\ddot {\bm \theta}+C\dot {\bm\theta}+{\bm g}({\bm \theta})+J^T\pazocal{F}_{\rm ext}=\bm \tau \text{,}
\label{sysdyn}
\end{equation}
where $\bm\theta \in \mathbb{R}^{6\times 1}$ is the joint angle vector, $J$ is the arm's Jacobian matrix, $M({\bm\theta})$, $C({\bm\theta},\dot{\bm \theta})$, and $\bm g(\bm \theta)$ are the mass matrix, Coriolis and centrifugal vector, gravity vector, respectively, and $\pazocal{F}_{\rm ext}=[\bm m^T_{\rm ext},\bm f^T_{\rm ext}]^T=[m_x, m_y, m_z, f_x, f_y, f_z]^T$ is the wrench applied by the end-effector to the environment. The control input to the robot is $\bm \tau$, the torque applied to the joint motors. 

The motion controller in the task space from \cite{murray1994mathematical} (pp. 195-197) is used. Define a smooth and invertible mapping from the joint space to task space $f: \bm \theta \in \mathbb{R}^{6\times 1} \to \bm x \in \mathbb{R}^{6\times 1}$. In this work, the joint space is mapped to the task space defined with the position and orientation of the end-effector, i.e., $ \bm x=[\alpha,\beta,\gamma,x,y,z]^T$. Note that this mapping is not globally but locally invertible thus works in a region, so multiple mappings may be used for the entire joint space. The Jacobian $J_f$ with respect to $f$ is defined as

\begin{equation}
\dot {\bm x}=J_f\dot {\bm \theta}  \text{,}\quad\quad  J_f=\frac{\partial f}{\partial {\bm \theta}} \text{.}
\end{equation}
Thus, 
\begin{equation}
\ddot {\bm x}=\dot J_f\dot {\bm \theta}+J_f\ddot {\bm \theta} \quad \Rightarrow \quad  \ddot {\bm \theta}=J_f^{-1}(\ddot {\bm x}-\dot J_f\dot {\bm \theta}) \text{.}
\label{fQ}
\end{equation}
The motion controller can be expressed as
\begin{equation}
\bm \tau=C\dot {\bm\theta}+{\bm g}({\bm\theta})+MJ_f^{-1}(\ddot {\bm x}_d+K_v\dot {\bm x}_e+K_p{\bm x}_e+K_i\int_0^t {\bm x}_eds-\dot J_f\dot {\bm\theta})\textbf{,}
\label{motionController}
\end{equation}
where ${\bm x}_e={\bm x}_d-{\bm x}$; ${\bm x}_d$ and ${\bm x}$ are the desired and actual configuration, respectively; $K_v$, $K_p$, and $K_i$ are the controller parameters and should be chosen properly to ensure stability \cite{murray1994mathematical} (pp. 198).

To introduce the hybrid controller, consider a task of cleaning a table with a robotic arm. The table surface is a plane parallel to the $x\text{-}y$ plane in world frame with the $z$ axis pointing upward.  Therefore, for such a task, a hybrid controller is needed to manipulate the end-effector to move from point $A$ to point $B$ on the table while maintaining a downward force (i.e., in -$z$ direction).

Define the mapping $f: {\bm\theta}  \to {\bm x}=[\alpha(t),\beta(t),\gamma(t),x(t),y(t),z(t)]^T $, where $x$, $y$, and $z$ are coordinates in world frame and the rest DOFs are orientation angles. We can decouple the 6 DOFs in the task space, i.e., the motion controller will be used to regulate $\{x(t),y(t),\alpha(t), \beta(t), \gamma(t)\}$ to track the reference trajectory and the force controller regulates the contact force in $z$ direction through manipulating $z(t)$. Let ${\bm x}_d=[\alpha_d(t), \beta_d(t), \gamma_d(t),x_d(t),y_d(t), z_{d0}]^T$ be the reference trajectory in the task space with $z_{r0}$ the initial $z$ position of the tip.

Plug Eqs.(\ref{motionController}) and (\ref{fQ}) into Eq. (\ref{sysdyn}), we can obtain
\begin{equation}
\ddot {\bm x}_e+K_v\dot {\bm x}_e+K_p{\bm x}_e+K_i\int_0^t {\bm x}_eds-J_fM^{-1}J^T\pazocal{F}_{\rm ext}=0
\label{errordyn}\text{.}
\end{equation}
Let $[\cdot]_{i}$ denote the $i$th elements of a vector and $u=J_fM^{-1}J^T\pazocal{F}_{\rm ext}$. Eq. (\ref{errordyn}) can be decomposed into two parts as follows.

\begin{equation}
    {[\ddot {\bm x}_e]}_{i}+K_v{[\dot {\bm x}_e]}_{i}+K_p{[{\bm x}_e]}_{i}+
K_i\int_0^t {[{\bm x}_e]}_{i}ds- 
[u]_{i}=0\text{,}  1\leq i \leq 5
\label{controller1}
\end{equation}

\begin{equation}
    [\ddot {\bm x}_e]_{6}+K_v {[\dot {\bm x}_e]}_{6}+K_p{[{\bm x}_e]}_{6}+
    K_i\int_0^t {[{\bm x}_e]}_{6}ds- 
    [u]_{6}=0 
 \label{controller2}
 \end{equation}

Eq. (\ref{controller1}) represents the motion error dynamics while Eq. (\ref{controller2}) will be used for the force controller design. Note that $\pazocal{F}_{\rm ext}$ is not canceled in Eq. (\ref{errordyn}) since it is difficult to accurately measure the term $\pazocal{F}_{\rm ext}$ and there will be a delay in canceling the term resulting in some unexpected behaviors. 

\textbf{Remark 1.} The force term $\pazocal{F}_{\rm ext}$ can be canceled (then $u=0$) if accurate sensors are available and the conclusion will not be affected. We are going to discuss the effect if this term is not canceled or just partially canceled.

\subsection{Hybrid Control: Part \Romannum{1}-Motion Control}
Eq. (\ref{controller1}) is the motion error dynamics after applying the motion controller Eq. (\ref{motionController}). Here we explain how the term $\pazocal{F}_{\rm ext}$ affects the controller performance. If $\pazocal{F}_{\rm ext}$ is not canceled then $[u]_i=[J_fM^{-1}J^T\pazocal{F}_{\rm ext}]_i$; if $\pazocal{F}_{\rm ext}$ is not fully canceled, $[u]_i$ represents the residual term after canceling, then we can obtain the transfer function of the system with input $u_i(t)$ and output $y_i(t)=[X_e]_i$ from Eq. (\ref{controller1}) as
\begin{equation}
G_i(s)=\frac{Y_i(s)}{U_i(s)}=\frac{s}{s^3+K_vs^2+K_ps+K_i} \text{.}
\label{lpfexp}
\end{equation}

If $K_v$, $K_p$, and $K_i$ are chosen such that Eq. (\ref{lpfexp}) is stable, it yields a low pass filter which has a very small gain in low frequency region. Thus, $u_i$ does not affect the tracking error $y_i$ much. In particular, when $s\to 0$, $G_i(s)\to 0$ which implies that constant $u_i$ cannot affect the output at all.

%For example, in the cases studies, we chose $K_v=35$, $K_p=405$, and $K_i=1500$, the bode plot of $G_i(s)$ is shown in Fig. \ref{fig2}. 

%%\begin{figure}[tb!]
%%\center
%%\includegraphics[trim={0cm 0cm 0cm 0cm},scale=0.5]{fig2.eps}
%%\vspace{-1em}
%%\caption{Bode plot of $G(s)$ for $K_v=35$, $K_p=405$, and $K_i=1500$.}
%%\label{fig2}
%%\vspace{1em}
%%\end{figure}

\subsection{Hybrid Control: Part \Romannum{2}-Force Control}
Let ${ z}_e={ z}_d-{ z}_t={[{\bm x}_e]}_{6}$ where ${ z}_d$ is the desired $z$ position and $z_t$ is the actual $z$ position of the end-effector. Let $f_e=f_d-f_{z}$, with $f_d$ the desired downward force and $f_{z}$ the actual downward force. We integrate another controller Eq. (\ref{controller2_2}) with Eq. (\ref{controller2}) as follows (Note that (\ref{controller2_1}) is obtained by replacing ${[\bm x_e]}_{6}$ with ${ z}_e$ in (\ref{controller2})).

\begin{empheq}[left={\empheqlbrace}]{alignat=2}
    &\ddot { z}_e+K_v\dot { z}_e+K_p{ z}_e+K_i\int_0^t { z}_eds-(\epsilon f_z+\eta)=0 
       \label{controller2_1}\\
    &u_c=\dot { z}_d=K_{p1}f_e+K_{i1}\int_0^t f_e ds
    \label{controller2_2} \text{~,}
\end{empheq}
where $(\epsilon f_z+\eta)=[J_fM^{-1}J^T\pazocal{F}_{\rm ext}]_{6}$, $u_c$ is the input for the force controller. In Eq. (\ref{controller2_1}), $f_z$ is extracted and the rest terms in $[J_fM^{-1}J^T\pazocal{F}_{\rm ext}]_{6}$ are accounted for with $\epsilon$ and $\eta$. Again, if $\pazocal{F}_{\rm ext}$ can be canceled, $\epsilon=0$, $\eta=0$. The reason why $f_z$ is extracted here is that $f_z$ can be very large thus dominates over other terms in $\pazocal{F}_{\rm ext}$ (i.e., $f_x$, $m_x$, etc.) when $f_d$ is large. 

\begin{figure}[tb!]
\center
\vspace{0.4em}
\includegraphics[trim={0cm 0cm 0cm 0cm},scale=0.7]{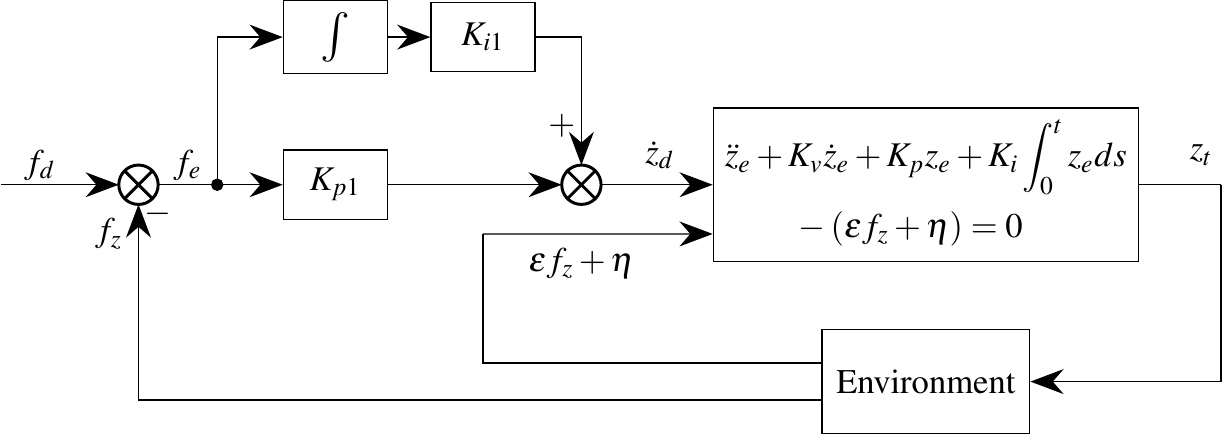}
\caption{Control diagram for the force controller.}
\label{block}
\vspace{1em}
\end{figure}

\begin{figure}[tb!]
\center
\includegraphics[trim={0cm 0cm 0cm 0cm},scale=1.5]{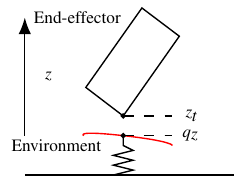}
\caption{Interaction model between the end-effector and the environment.}
\label{interModel}
\vspace{1em}
\end{figure}

Controller Eq. (\ref{controller2_2}) essentially indirectly regulates the contact force through adjusting the desired $z$ position. The block diagram of the force controller is shown in Fig. \ref{block}.
We use a spring model to approximate the interaction dynamics between the tip and the environment. As seen in Fig. \ref{interModel}, let $q_z$ be the $z$ position of the contact point before contacting (before contacting there is no deformation at the contact point), the contact force can be computed with $f_z=\delta(k)({q_z}-z_t)$, where 
\begin{equation}
\begin{aligned}
\delta(k)=
\begin{cases}
 &k  \text{~, if}~~q_z>z_t\\
 &0\text{~, if}~~q_z \leq z_t
\label{deltak}
\end{cases}
\end{aligned}\text{.}
\end{equation}

Sometimes the environment may be rigid and there is no deformation, the model can still work due to the flexibility of the robotic arm. When implementing the controller, $f_z$ can be read from the F/T sensor.

%\begin{equation}
%\begin{aligned}
% f_z=\phi(T_t,Z_t)=
%\begin{cases}
% &k(T_t-Z_t) \text{ if}~~T_t>Z_t\\
% &0\text{ ~~~~\quad \quad   if}~~T_t \leq Z_t
%\label{spring1}
%\end{cases}
%\end{aligned}
%\end{equation}

To obtain the closed-loop force error dynamics, the system state $W=[w_1, \cdots, w_6]^T$ is defined as follows. For simplicity, it is assumed  that $\dot f_d=0$, but this assumption can be removed and similar analysis can be carried out.

\begin{equation}
\begin{aligned}
\begin{cases}
w_1&=\int_0^t z_e ds\\
w_2&=z_t\\
w_3&=\dot z_e\\
w_4&=z_d\\
w_5&=\int_0^t f_e ds\\
w_6&=f_e
\end{cases}
\end{aligned}\Rightarrow
\begin{aligned}
\begin{cases}
\dot w_1&=z_e=z_d-z_t=w_4-w_2\\
\dot w_2&=\dot z_t=\dot z_d-\dot z_e=u_c-w_3\\
&=-w_3+K_{i1}w_5+K_{p1}w_6\\
\dot w_3&=\ddot z_e=-K_vw_3-K_p(w_4-w_2)\\
&~~-K_iw_1+(\epsilon f_z+\eta)~ (\text{by Eq. (\ref{controller2_1})})\\
\dot w_4&=\dot z_d=u_c=K_{i1}w_5+K_{p1}w_6\\
\dot w_5&= f_e =w_6\\
\dot w_6&=\dot f_e=-\dot f_z
\end{cases}
\end{aligned}
\label{ssforcecontroller}
\end{equation}

%With Eq. (\ref{deltak}), it follows that $f_z=\delta(k)(T_t-Z_t)=\delta(k)(T_t-w_2)$ and $\dot f_z=\delta(k)(\dot T_t-\dot Z_t)=\delta(k)(\dot T_t-u_c+w_3)$. Define state vector $W=[w_1,\cdots,w_6]^T$. Therefore, Eq. (\ref{ssforcecontroller}) can be written in state-space form as follows.
The system state-space equation is

\begin{equation}
\hspace*{-1em}
\begin{aligned}
\dot W&=A_wW+BU_w\\
=&\begin{bmatrix}
0&-1&0&1&0&0\\
0&0&-1&0&K_{i1}&K_{p1}\\
-K_i&\epsilon\delta(k)+K_p&-K_d&-K_p&0&0\\
0&0&0&0&K_{i1}&K_{p1}\\
0&0&0&0&0&1\\
0&0&-\delta(k)&0&\delta(k)K_{i1}&\delta(k)K_{p1}
\end{bmatrix}W\\
+&
\begin{bmatrix}
0&0\\
0&0\\
1&0\\
0&0\\
0&0\\
0&1
\end{bmatrix}
\begin{bmatrix}
\eta-\epsilon \delta(k)q_z\\
-\delta(k)\dot q_z
\end{bmatrix}
\end{aligned}\text{.}
\label{ssmodel}
\end{equation}

Next, we analyze the case when $\epsilon(t)=0$ and $\eta(t)=0$, i.e., $\pazocal{F}_{\rm ext}$ can be canceled. If $\epsilon(t)\neq0$, the system Eq. (\ref{ssmodel}) will be a time-varying model which is more challenging to analyze its closed-loop stability.

\textbf{Non-contact Mode:} If $q_z\leq z_t$ for $t \geq t_0$, there is no contact, thus $ w_6=0$, $f_e=f_d$,  the controller Eq. (\ref{controller2_2}) becomes
\begin{equation}
\begin{aligned}
 u_c(t)&=K_{p1}f_d+K_{i1}\int_0^{t_0}f_eds+K_{i1}f_d(t-t_0)\\
 &=\dot z_d(t_0)+K_{i1}f_d \cdot (t-t_0) \text{,}
\label{ut}
\end{aligned}
\end{equation}
where $t\geq t_0$. Eq. (\ref{ssmodel}) degenerates to
\begin{equation}
\dot {[W]}_{1:3}=\begin{bmatrix}
0&-1&0\\
0&0&-1\\
-K_i&+K_p&-K_d
\end{bmatrix}[W]_{1:3}+
\begin{bmatrix}
 z_d(t_0)+\int_{t_0}^t u_c(t)ds\\
 u_c(t)\\
0
\end{bmatrix}
\label{ssmodel2}\text{.}
\end{equation} 

As seen in Eq. (\ref{ut}), $u_c(t)$ will monotonously decrease or increase, which may result in very large $u_c(t)$ before reaching the surface again if $\|z_d-q_z\|$ is very large. To avoid this and make the controller more robust, we can add a saturation function to the controller. Specifically, assume the estimated contact point position $\hat{q}_z$ satisfies $-\Delta<\hat{q}_z-q_z<0$ ($\Delta>0$ can be seen as the estimation error bound), the controller Eq. (\ref{controller2_2}) is modified to

\begin{equation}
\begin{aligned}
u_c=
\begin{cases}
 &S(K_{p1}f_e+K_{i1}\int_0^t f_e ds)  \text{~~if}~~f_z=0\\
 &K_{p1}f_e+K_{i1}\int_0^t f_e ds  \text{~~~~~~if}~~f_z\neq 0
\end{cases}
\end{aligned} 
\label{modifiedcontroller}
\end{equation}
with
\begin{equation}
\begin{aligned}
S(z)=
\begin{cases}
 &0 \text{~~if}~~z_d<\hat{q}_z\\
 &z \text{~~otherwise}
\end{cases}\text{.}
\end{aligned}
\end{equation}

Since $K_p$, $K_v$ and $K_i$ can be chosen to make Eq. (\ref{ssmodel2}) stable, the end-effector will eventually reach the surface again, then the controller will switch to contact mode as explained next.

\textbf{Contact Mode:} If $q_z>z_t$, choose $K_{i1}$ and $K_{p1}$ such that all the non-zero eigenvalues of $A_w$ have negative real parts. However, since $\text{rank}(A_w)=5$, there will be a zero eigenvalue for $A_w$. We have the following proposition.

\textbf{Proposition 1.} In contact mode, suppose $\pazocal{F}_{\rm ext}$ can be canceled, i.e., $\epsilon=0$ and $\eta=0$, choose $K_{i1}$ and $K_{p1}$ such that all the non-zero eigenvalues have negative real parts, then,
\begin{enumerate}
\item if $q_z=0$ and $\dot q_z=0$, $f_e$ converges to 0, i.e., for the system $\dot W=A_wW$, $\lim\limits_{t\to \infty}f_e=\lim\limits_{t\to \infty}w_6=0$;
\item if $q_z\neq 0$ and $\dot q_z\neq 0$, $\lim\limits_{t\to \infty}\|f_e\|=\lim\limits_{t\to \infty}\|w_6\|\leq \Gamma \|U_w\|_{\infty} $, where $\Gamma$ is a constant.
\end{enumerate}

\textbf{Proof.} \textbf{1)} Here we consider the case in which $A_w$ has distinct eigenvalues, if $A_w$ has repeated eigenvalues, as long as they have negative real parts, the similar analysis can be performed based on linear system theory without changing the conclusion \cite{chen1970introduction}. 

The solution of $\dot W=A_w W$ is $W(t)=\sum\limits_{i}c_ie^{\lambda_i t}{\bm v}_i$ ($c_i$ is a constant scalar depending on the initial state, ${\bm v}_i$ is the eigenvector corresponding to the eigenvalue $\lambda_i$), $\lim_{t\to \infty}c_ie^{\lambda_i t}{\bm v}_i = 0$ if $\text{real}(\lambda_i)<0$. Since there is only one zero eigenvalue $\lambda_6$, thus $\lim_{t\to \infty}W(t)=c_6{\bm v}_6$. The eigenvector corresponding to the zero eigenvalue is ${\bm v}_6$, then $A_w{\bm v}_6=0$, we have
\begin{equation}
\begin{aligned}
\begin{cases}
-[{\bm v}_6]_3+K_{i1}[{\bm v}_6]_5+K_{p1}[{\bm v}_6]_6=0\\
K_{i1}[{\bm v}_6]_5+K_{p1}[{\bm v}_6]_6=0\\
[{\bm v}_6]_6=0
\end{cases}
\end{aligned}\Rightarrow
\begin{aligned}
\begin{cases}
[{\bm v}_6]_3=0\\
[{\bm v}_6]_5=0\\
[{\bm v}_6]_6=0
\end{cases}
\end{aligned}\text{.}
\label{Winfty}
\end{equation}
It follows that $\lim\limits_{t\to \infty}f_e=\lim\limits_{t\to \infty}w_6=c_6[{\bm v}_6]_6=0$, implying that the system is asymptotically stable in contact mode. Eq. (\ref{Winfty}) also indicates that $\dot {z}_e \to 0$ and $\int_0^{t} f_e ds\to 0$ as $t\to \infty$. \hspace*{\fill}$\blacksquare$

\textbf{2)} Let $y=f_e=CW=[0, 0, 0, 0, 0, 1]W$ be the output of the system. Then, the state-space model can be expressed as
\begin{equation}
\begin{aligned}
\dot W&=A_wW+BU_w\\
y(t)&=CW
\end{aligned}\textbf{.}
\label{ssmodel3}
\end{equation}
The solution of Eq. (\ref{ssmodel3}) is known as \cite{chen1970introduction}
\begin{equation}
y(t)=Ce^{A_wt}W(0)+C\int_0^te^{A_w(t-\tau)}BU_w(\tau)d\tau \text{,}
\end{equation}
where $W(0)$ denotes the state at $t=0$. As $A_w$ has distinct eigenvalues, $A_w$ can be decomposed as $A_w=VDV^{-1}$, where $D=\text{diag}\{\lambda_1,\lambda_2,\lambda_3,\lambda_4,\lambda_5,0\}$ and real$(\lambda_i)<0$, the $i$th column of $V$ is an eigenvector of $A_w$ corresponding to eigenvalue $\lambda_i$. Therefore, $e^{A_wt}=Ve^{Dt}V^{-1}$, $e^{Dt}=\text{diag}\{e^{\lambda_1t},\cdots,e^{\lambda_5t},1\}$. Let ${\bm \gamma}_6$ denote the 6th row of $V$, then
\begin{equation*}
\begin{aligned}
\lim\limits_{t\to \infty}Ce^{A_wt}W(0)&=\lim\limits_{t\to \infty}CV\text{diag}\{e^{\lambda_1t},\cdots,e^{\lambda_5t},1\}V^{-1}W(0)\\
&={\bm \gamma}_6\text{diag}\{0,0,0,0,0,1\}V^{-1}W(0)=0\text{,}
\end{aligned}
\end{equation*}
where we have used the fact that $[{\bm \gamma}_6]_6=0$ which is proven by Eq. (\ref{Winfty}). Let $V^{-1}=[{\bm \alpha}_1,\cdots, {\bm \alpha}_6]$, where ${\bm \alpha}_i$ is a column vector, we have
\begin{equation*}
\begin{aligned}
&\int_0^tCe^{A_w(t-\tau)}BU_w(\tau)d\tau\\
&=\int_0^tCV\text{diag}\{e^{\lambda_1(t-\tau)},\cdots,e^{\lambda_5(t-\tau)},1\}V^{-1}BU_w(\tau)d\tau\\
&=\int_0^t {\bm \gamma}_6\text{diag}\{e^{\lambda_1(t-\tau)},\cdots,e^{\lambda_5(t-\tau)},1\}[{\bm \alpha}_3,{\bm \alpha}_6]U_w(\tau)d\tau\\
&=\int_0^t[ [{\bm \gamma}_6]_1e^{\lambda_1(t-\tau)},\cdots,[\bm \gamma_6]_5e^{\lambda_5(t-\tau)},0][{\bm \alpha}_3,{\bm \alpha}_6]U_w(\tau)d\tau
\end{aligned}\text{.}
\end{equation*}
Since $\lim\limits_{t\to \infty}\int_0^{t}e^{\lambda_i(t-\tau)}d\tau<\infty$ for $\lambda_i<0$, there exists a constant $\epsilon_i>0$ such as $\lim\limits_{t\to \infty}\|\int_0^{t}e^{\lambda_i(t-\tau)}d\tau\|<\epsilon_i$. Therefore, 
\begin{equation*}
\begin{aligned}
&\lim\limits_{t\to \infty}\|f_e\|=\lim\limits_{t\to \infty}\|y(t)\|\\
&\leq\norm{\int_0^t[ [\bm \gamma_6]_1e^{\lambda_1(t-\tau)},\cdots,[\bm \gamma_6]_5e^{\lambda_5(t-\tau)},0][\bm \alpha_3,\bm \alpha_6]U_w(\tau)d\tau}\\
&\leq [|[\bm\gamma_6]_1|\epsilon_1,\cdots,|[\bm\gamma_6]_5|\epsilon_5,0][|\bm \alpha_3|,|\bm \alpha_6|][1,1]^T\cdot \|U_w\|_{\infty}\\
&=\Gamma \|U_w\|_{\infty}
\end{aligned}\text{,}
\end{equation*}
where $\gamma$ is a constant, and $|\cdot|$ on a vector is element-wise, i.e., taking absolute value for each entry of a vector.

%&\leq \norm{\int_0^t[[\bm\gamma_6]_1e^{\lambda_1(t-\tau)},\cdots,[\bm \gamma_6]_5e^{\lambda_5(t-\tau)},0][\bm \alpha_3,\bm \alpha_6][1,1]^Td\tau \|U_w\|_{\infty}}\\

If $A_w$ has repeated eigenvalues, $D$ will be a Jordan matrix and similar arguments can be used to prove \textbf{2)} with the linear system theory \cite{chen1970introduction}.\hspace*{\fill} $\blacksquare$

\textbf{Remark 2.} \textbf{2)} also implies that the system is bounded-input-bounded-output (BIBO)\cite{khalil2002nonlinear}. 

\textbf{Hybrid Mode:} Usually, the system will start from non-contact mode and remain in contact mode, but it is also possible that it switches between the two modes when $q_z$ is changing. In such case, it is difficult to show the asymptotically stability, however, we can show that the system output is bounded with the proposed controller applied: if the system stays in contact mode for infinite time, the system is asymptotically stable thus bounded based on the above discussion; if it switches to non-contact mode at $t_0$, then the output at $t_0$ must be a finite value thus bounded, once entering non-contact mode the output is still bounded due to the saturation function in the modified controller (Eq. (\ref{modifiedcontroller})). Therefore, the output $z_t$ is always bounded with the proposed force controller.

\textbf{Remark 3.} Eq. (\ref{controller2_2}) is a PI controller, it can also be upgraded to a PID controller as $u_c=\ddot z_d=K_{v1}\dot f_e+K_{p1}f_e+K_{i1}\int_0^t f_e ds$. The analysis can be carried out in the same way by adding an extra state $w_7=\dot z_d$ to Eq. (\ref{ssforcecontroller}). Note that this analysis is quite conservative. In simulation (e.g., in case study A, the topography $q_z$ is changing), it turns out that $\lim_{t\to \infty}f_e=0$ when $\dot {q}_z$ is not large.

\textbf{Remark 4.} In Eq. (\ref{ssforcecontroller}), $A_w$ is rank-deficient due to the linear interaction model. The rank deficiency is not guaranteed if other interaction models were used. 

\textbf{Remark 5.} In non-contact mode, extra information is needed to make sure that end-effector can reach the object surface, i.e., entering contact-mode again. In this case, this extra information is that by moving in -$z$ direction, it is guaranteed to reach the surface. Note that the signs of $K_{p1}$ and $K_{i1}$ are determined with such extra information.

\section{Case Studies}
\subsection{Contour Tracking/Polishing Parts}
\begin{figure}[tb!]
\center
\vspace{0.5em}
\includegraphics[trim={0cm 0cm 0cm 0cm},scale=1]{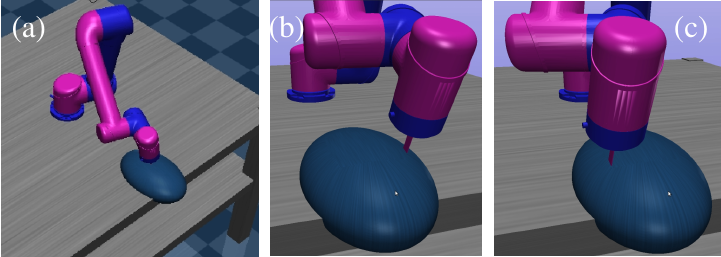}
\caption{(a) Setup for Case A. (b) Initial state. (c) Intermediate state during contour tracking.}
\label{expmujoco2_1}
\vspace{1em}
\end{figure}

This case is similar to the task presented in \cite{schindlbeck2015unified}. The goal is to explore everywhere on the object surface (position control) with constant downward force (force control). We can directly apply the hybrid controller here with force in $z$ direction maintained. The process is also equivalent to contour tracking with constant force as shown in Fig. \ref{expmujoco2_1}. In the simulation the object is an ellipsoid as shown in Fig. \ref{expmujoco2_1}.

\subsection{Grabbing a Box with Two Robotic Arms}
\begin{figure}[t!]
\centering
\vspace{0.2em}
\includegraphics[scale=1.3]{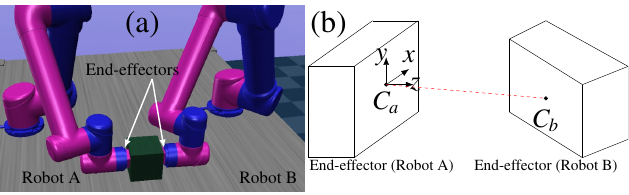}
\caption{Case B: grabbing with two robotic arms.}
\label{grabbing}
\vspace{1em}
\end{figure}

Using two arms to grab a box is a common task for human beings. To mimic this with robots, consider the scenario in which two robotic arms are used to grab a box as shown in Fig. \ref{grabbing} (a). 

During the process, the end-effector of two robots have to track the pre-designed trajectories, and maintain a force in the $z$ direction of the frame with origin at $C_a$ as shown in Fig. \ref{grabbing} (b). Therefore a motion/force controller is needed. The proposed hybrid controller can be applied in this case through modifying the mapping $f$ in Eq. (\ref{fQ}). Note that the hybrid controller will only be applied to robot B as shown in Fig. \ref{grabbing} (a) in order to establish an invertible and smooth mapping from the joint space to the task space as explained next. Robot A is controlled with a motion controller. 

As seen in Fig. \ref{grabbing} (a), the frame ($C_a$ is the frame origin) attached to the end-effector of robot A can be determined with $\phi(t)=\{(x_{a}(t), y_{a}(t), z_{a}(t)),[V_x(t), V_y(t), V_z(t)]\}$, where $(x_{a}(t), y_{a}(t), z_{a}(t))$ is the coordinate of $C_a$ in world frame and $[V_x(t), V_y(t), V_z(t)]^T \in  \mathbb{R}^{3\times 1}$ represents the directions of the $x$, $y$, and $z$ axes of the attached frame in world frame. Note that once the trajectory of end-effector of robot A is given, $\phi(t)$ is fixed at time $t$ assuming the motion controller works effectively. Recall that the aforementioned mapping $f: {\bm\theta} \in Q \to \bm x \in \mathbb{R}^{6\times 1}$ with $\bm x=[\alpha, \beta, \gamma, x, y, z]^T$. Now for the end-effector of robot B, define a new mapping $h: \bm x \to \bar{\bm x}$ where $\bar{\bm x}=[\alpha, \beta, \gamma, \bar{x}, \bar{y}, \bar{z}]^T$ and $\bar{x},\bar{y},\bar{z}$ can be computed as follows.

\begin{equation}
\begin{bmatrix}
\bar{x}\\
\bar{y}\\
\bar{z}
\end{bmatrix}=\begin{bmatrix}
V_x^T\\
V_y^T\\
V_z^T
\end{bmatrix}\Bigg(
\begin{bmatrix}
x\\
y\\
z
\end{bmatrix}-
\begin{bmatrix}
x_a\\
y_a\\
z_a
\end{bmatrix}\Bigg)
\label{mappingtask3}
\end{equation}

It can be verified that $h$ is a smooth and invertible mapping. Therefore, $h\circ f$, the composition of two smooth and invertible mappings, is still a smooth and invertible mapping. Essentially, $h$ maps the coordinate of $C_b$ to the frame originated at $C_a$. Thus the force in the $z$ direction of frame $C_a$ is determined by $\bar{z}$ which is the same with the hybrid controller we discussed before. Therefore, the same hybrid controller can be applied by modifying the mapping from $f$ to $h\circ f$. 

Note that this does not mean that the box is guaranteed to be picked up and follow the desired trajectory, the force-closure conditions are required to hold \cite{lynch2017modern}. Here we assume that the force-closure conditions hold by adjusting properly the initial orientation and position of the two end-effectors as shown in Fig. \ref{grabbing} (a).

\begin{figure*}[h!]
\center
\includegraphics[trim={2cm 0cm 0cm 0cm},scale=0.59]{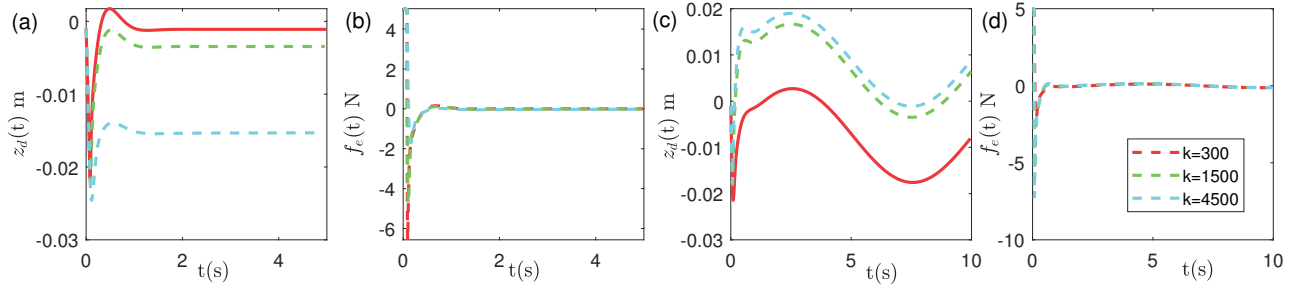}
\caption{(a) and (c) Desired positions during control for constant  and varying $q_z$, respectively. (b) and (d) Force errors for constant  and varying $q_z$, respectively.}
\label{ForceSim}
\vspace{-1.2em}
\end{figure*}

\begin{figure*}[t!]
\center
\includegraphics[trim={3.5cm 0cm 0cm 0cm},scale=0.6]{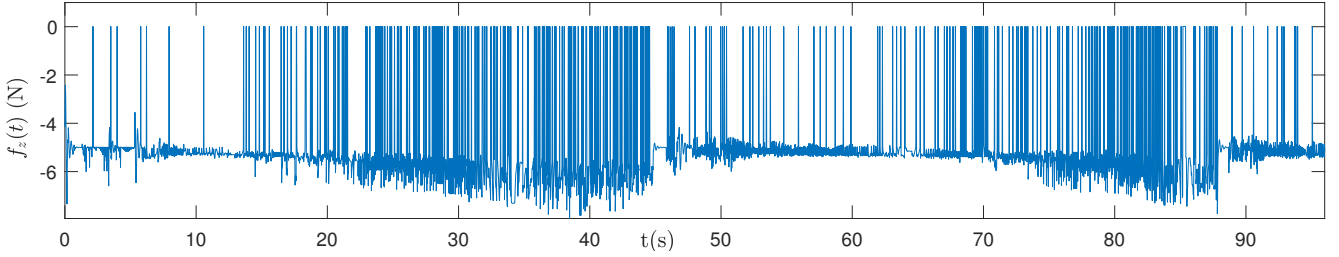}
\caption{The force profile in $z$ direction for case A.}
\label{expfig1}
\vspace{-1.2em}
\end{figure*}

\begin{figure}[t!]
\center
\includegraphics[trim={0.5cm 0cm 0cm 0cm},scale=0.52]{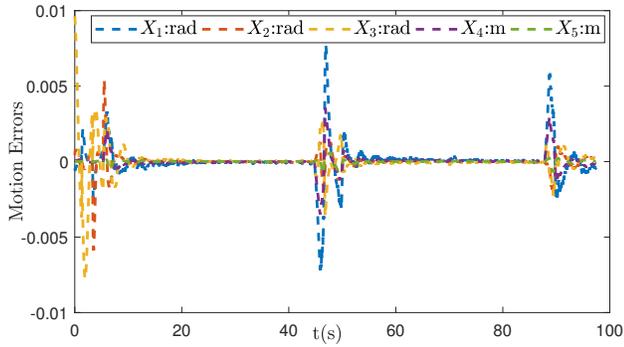}
\caption{Position control errors for case B.}
\label{expfig2}
\vspace{0.5em}
\end{figure}

\section{Simulation Results}
\subsection{Force Controller Simulation Results}
To simulate the force controller Eq. (\ref{controller2_1})-(\ref{controller2_2}), parameters of controller Eq. (\ref{controller1}) were set as $K_v=35$, $K_p=405$, and $K_i=1500$. The desired force was $f_z=5$N. The PI force controller parameters are $K_{p1}=-0.05$ and $K_{i1}=-0.01$. We simulated the contact model for three different spring constants: $k=300$N/m, $k=1500$N/m, and $k=4500$N/m. At the beginning, the tip position was $z_t=0.05$m. The initial desired position was $z_d=-0.001$m. 

If $\epsilon=0$ and $\eta=0$, it can be checked that all the nonzero eigenvalues of $A_w$ have negative real parts, thus the above analysis can be applied. Fig. \ref{ForceSim} (a) and (b) show the performance of the force controller for different $k$s when $q_z=0$. It can be seen that the system starts from non-contact mode and then switches to contact model and remains there for the rest time. The force can be accurately regulated for different $k$s and converges in about 0.1s.

To test the controller performance for time-varying $q_z$, we simulated the controller with $q_z(t)=0.01\sin(2\pi0.2t)+0.01$m. The results are shown in Fig. \ref{ForceSim}(c) and (d). This can happen when the tip is sliding on a curved surface. It can be seen that the converged force error is very small for different $k$ even when the contact point changes from 0.02m to 0m in about 5s.

The robotic arm used is UR5 which has the size of a human arm and the simulation platform is MuJoCo \cite{todorov2014convex}. For both two cases, the controller parameters are $K_v=35$, $K_p=405$,  $K_i=1500$, $K_{p1}=0.008$ and $K_{i1}=0.002$. The video is available in the supplementary material.

\subsection{Case A: Contour Tracking/Polishing Parts}
In this case, the topography $q_z$ is changing as shown in Fig. \ref{expmujoco2_1}. The force curve during the polishing process is shown in Fig. \ref{expfig1} (b) and the downward force is maintained to $-5$N.

\subsection{Case B: Grabbing with Two Robotic Arms}
\begin{figure}[t!]
\center
\includegraphics[trim={1cm 0cm 0cm 0cm},scale=0.5]{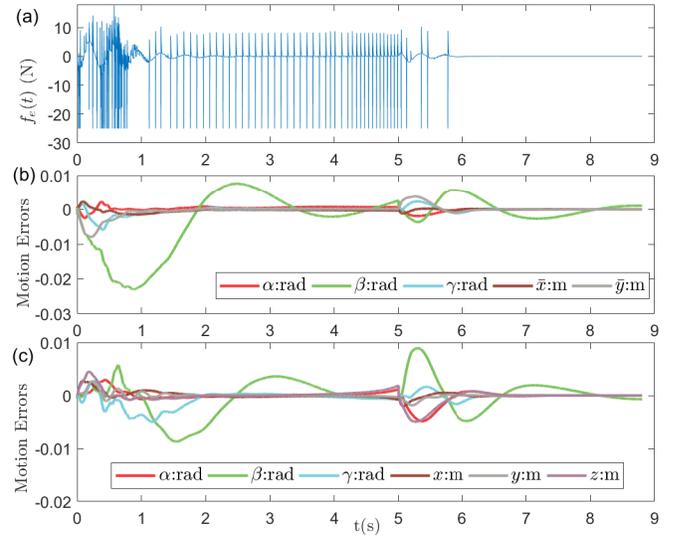}
\vspace{-2em}
\caption{(a) Force control errors for case B. Motion errors for (b) robotic arm A and (c) robotic arm B.}
\label{expfig3}
\vspace{1em}
\end{figure}

In this case, the goal is to lift the box with mass of 1kg and move along the vector $[0.27, 0.1,0.27]$m within 5s in the world frame. Note that the hybrid controller is only applied to robotic arm B while the other one is controlled by a motion controller alone. The force is regulated to 20N with the friction coefficient set to be 0.9 and the force control error is shown in Fig. \ref{expfig3} (a). The motion control errors for the two robots are shown in Fig. \ref{expfig3} (b) and (c), respectively. Note that for robotic arm B the motion error are expressed in term of $\bar x$ and $\bar y$ based on the mapping in Eq. (\ref{mappingtask3}). Again, the hybrid controller works as expected with both the motion and force regulated.

Note that for both cases the force error is very noisy as can be seen in Figs. \ref{expfig1} and \ref{expfig3} (a). This is partly due to the fact that the environment in simulation is very hard. The contact is shortly broken and then established again due to the controller resulting in spikes in the plots.  On the other hand, the algorithm for calculating the interaction force in MuJoCo may also contribute to this. Computing the interaction force is transformed to solving an optimization problem in MuJoCo making it very challenging to set parameters to change the elasticity of the environment.

\section{Conclusion and Future Work}
In this paper, a hybrid position/force controller is proposed. The position and force controllers correspond to different DOFs in the task space with the smooth and invertible mapping. The simulation results validated the proposed approach.

The interaction model used in this work is linear, which may not be the real case, more complex interaction models will be considered in the future. More examples in which the force controller corresponds to more DOFs will be studied. In addition, experiments with real robots will be conducted to verify the proposed method.

Although we have not observed Zeno behavior of the hybrid controller in simulations, current analysis cannot rule out the possibility of this  phenomena which needs further investigations.

\section*{ACKNOWLEDGMENT}
Shengwen would like to thank Professor Yan-Bin Jia for the technical discussions which improved the quality of this work.
%The financial support from National Science Foundation CMMI-1634592 is gratefully acknowledged.

\bibliography{hybrid}
\bibliographystyle{ieeetr} 

\end{document}